# Agile Satellite Planning for Multi-Payload Observations for Earth Science

**Rich Levinson**[1,2]  **Sreeja Nag**[1,3]  **Vinay Ravindra**[1,3]

[1]NASA Ames Research Center, Moffett Field, CA 94035
[2]KBR Wyle Services, LLC, Moffett Field, CA 94035
[3]Bay Area Environmental Research Institute, Moffett Field, CA 94035

**Abstract**

We present planning challenges, methods and preliminary results for a new model-based paradigm for earth observing systems in adaptive remote sensing. Our heuristically guided constraint optimization planner produces coordinated plans for multiple satellites, each with multiple instruments (payloads). The satellites are agile, meaning they can quickly maneuver to change viewing angles in response to rapidly changing phenomena. The planner operates in a closed-loop context, updating the plan as it receives regular sensor data and updated predictions. We describe the planner's search space and search procedure, and present preliminary experiment results. Contributions include initial identification of the planner's search space, constraints, heuristics, and performance metrics applied to a soil moisture monitoring scenario using spaceborne radars.

## Introduction

Agile and maneuverable satellites provide an opportunity for rapid and direct coupling between model updates and planning new observations to improve the models based on a model quality metric. NASA calls this new paradigm which extends the concept of Sensor Webs (Mandl et al., 2006) to model-driven observation planning, New Observing Strategies (NOS) (Le Moigne et al.) We present the planning component for one such NOS, called Distributed Spacecraft with Heuristic Intelligence to Enable Logistical Decisions, or D-SHIELD (Nag et. al. 2020, 2021).

Missions in the past have physically re-pointed single instruments given ground-commanded waypoints (CHRIS on Proba (Barnsley et al., 2004)), 3-DOF imaging for Planet's Skybox spacecraft (Augustein et al., 2016), and EO-1 re-tasking for monitoring of floods (Chien et al 2019), volcanoes (Chien et al 2020), and wildfires (Chien et al 2011). Missions without physical agility have also shown to benefit from reactive planning to prioritize hyperspectral data collection, such as IPEX which served as the HyspIRI pathfinder (Chien et al., 2016) and the future EnMAP (Worle et al 2014, Fruth et al 2019), and to inform operational parameters like electronic beam steering to optimize radar looks, such as TerraSAR-X (Werninghaus and Buckreuss 2009) and TanDEM-X (Krieger et al 2007). Power and bandwidth restrictions on small spacecraft has spurred literature on scheduling data download (Jian and Cheng, 2008) and use of crosslinks to propagate planning information via space nodes (Linnabary et al., 2019) however these tools are optimized only for data downlink without a science-driven observation scheduler in the loop.

D-SHIELD's planner may be the first to create coordinated plans for multiple physically agile satellites, each with multiple instruments (payloads) and explicit models of measurement errors applied to global soil moisture estimation, using an architecture that is flexible between on ground and onboard. The planner operates in a closed-loop context, updating the plan as it receives regular sensor updates from the rest of the constellation, other space assets (e.g. SMAP, CYGNSS satellites) or in-situ ground networks (e.g. SoilScape). While our previous work has showed science-in-the-loop planning for single sensors per satellite in a constellation (Nag et al., 2019), this application domain with multiple payloads heterogeneously spread in a constellation has not previously been defined in formal planning terms. We describe the planning challenges and solution, detailing the search space and search procedure, and present preliminary experiment results. Contributions include initial dentification of the planner's search space, constraints, heuristics, and performance metrics.

## Problem Description

D-SHIELD uses a (proposed) constellation of satellites looking at Earth to reduce global soil moisture uncertainty by making observations that target spatio-temporal points of rising prediction error, which accurate data can alleviate. Such uncertainty reduction benefits accurate models for floods, wildfires, vegetation drydowns, etc. The overall goal is to demonstrate a system which continually updates a hydrologic land surface model with external forcing functions (e.g. precipitation) and dynamically schedules new observations to improve the model where the error is greatest, such as right after rain occurs, or places which have not been observed recently, using instrument parameters that minimize retrieval errors. The dynamic soil moisture model detects regional model quality degradation in near real-time and provides input to the planner about the highest priority ground positions (GP) to observe next from a science perspective. This paper focuses on the D-SHIELD planner more than the science model.

Each satellite in the constellation includes at least 2 different radar instruments (L-band and P-band) to take images of ground positions (GP). Each image typically covers multiple GP. Each satellite has a set of access times when it can view various GP, based on its orbit. These access times are called timepoints (TP). The planner must decide which satellites look at which GPs, at what times, with which instruments using which available viewing angles.

An "observation" may involve a single instrument at a single TP or may combine both instruments and/or multiple observations of a GP at separate times to improve the quality of the soil moisture model. For any observation, soil moisture can be retrieved from the radar measurements and is expected to reduce the predictive uncertainty around the soil moisture of the observed GPs.

Explicit Error Model: D-SHIELD uses high-fidelity models of spacecraft and soil moisture dynamics, combined with a constraint satisfaction planner to generate new observation plans for all satellites in the constellation with the objective of improving model quality. Model quality is inversely proportional to the model error associated with each GP. Model error for each GP is a combination of a given prior model error (predictive uncertainty) and measurement error (soil moisture estimate retrieval error) from new observations. Model error increases over time without new observations and increases when rain occurs. Measurement error is a function of which instrument (L-band or P-band) is used, the viewing angle, the type of ground cover (e.g., barren, shrubs, forest, croplands), and some other ancillary parameters. For example, if a P-band measurement does not improve errors, the planner may decide to forego it and conserve battery power in an eclipse.

**Constraints:** The planner must enforce the constraints:

Image Lock - Each observation requires the instrument to hold its viewing angle for 3 seconds so that a stripmap image can be created. This blocks out slewing to another viewing angle during that 3-second image lock period.

Duplicate observations - We do not look at the same position twice in the same 24-hour period to cover more unobserved locations. There are exceptions for serendipitous cases when we aim at one GP but capture others in the same image, and for intentional cases when we plan a follow-up observation.

Maneuver constraints (slew time and energy) - The satellite must slew to change viewing angles. There are constraints on how quickly it can change viewing angles, depending on the slew magnitude. Changing viewing angle takes a different amount of time and energy depending on the combination of initial angle and target angle. This is called the slew time constraint. Energy consumption is also dependent on the slew magnitude. The planner must ensure there is enough time to slew between each observation and track the energy consumed by each slew to enforce a minimum energy level constraint.

Energy model and constraints - The planner must track energy models for each satellite independently to ensure energy remains above a 70% minimum charge level to preserve battery health. Energy is consumed at a steady rate whenever an instrument is on (taking images) plus a variable amount of energy is consumed with each slew. The charge level is increased at a steady rate by solar panels, except during eclipse when the charge is not increased.

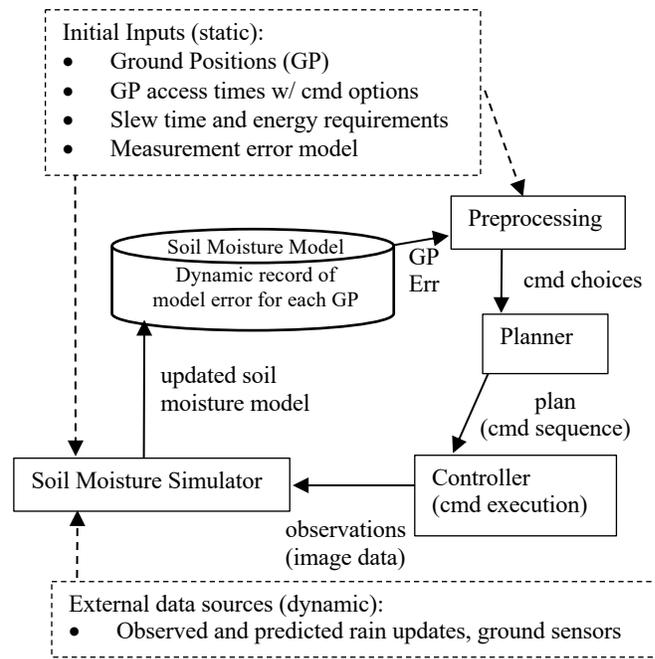

**Figure 1: D-SHIELD architecture**

Figure 1 shows the D-SHIELD system. The solid lines show the closed-loop control flow while the dashed lines show external data sources used to update the model. The process is initialized with inputs based on the satellite orbits and specifications for the spacecraft and instruments, which are used to determine available observation times, along with slew time and energy requirements. Those raw inputs go through *pre-processing* to generate the planner input files. The planner searches through the space of all available observation times and decides what to look at, when to look at it, and how to look at it. The planner produces a sequence of commands for the Controller to execute (command the instruments to take the images).

The (simulated) Controller collects the observation data and passes that to the Soil Moisture Simulator, to update the model based on the new observations. The Soil Moisture Model is a dynamic database which maintains a record of the current model error for each GP.

**The planner's job** is to create a coordinated multi-satellite observation plan which improves the model quality by observing the GP with the highest model error using measurements with the least error. Each plan step is an observation command of the form <time, instrument(s), viewing angle>, which specifies the time when one or both instruments will take images at the given viewing angle.

**Measurement Error Table:** Planner input includes an error table which defines the expected measurement error for any combination of instruments and viewing angles, which the planner minimizes when choosing commands.

**Follow-up observations.** The planner may choose to have a satellite take multiple images of same GP to minimize error. Examples include:

- Same instrument at different times (at the same or different angles)
- Both instruments at the same time (at the same angle)
- Both instruments at different times (may involve different orderings of the instruments)

Choices involving a follow-up observation within 2 hours are considered a single observation by the science model because it can combine several observations within up to 2 hours to reduce speckle noise in the soil moisture estimation retrieval. Scheduling a follow-up means sometimes a choice about what to do at one time requires reserving a timeslot in the future, placing constraints on a future TP. In these cases when we explicitly want to observe a GP twice at different times, the planner must choose the ordering for the viewing angles. This is an exception to our default constraint of no duplicate observations, driven by the measurement error table indicating that the follow-up image will significantly reduce error.

**Continual re-planning:** D-SHIELD is designed to run iteratively and continually. Each planning iteration generates a plan for 6 hours (configurable), after which the plan is executed, producing new measurements. New observations are passed to the soil moisture simulator (Nag et al., 2021) which assimilates the data and updates predictions of soil moisture uncertainty across all GPs in the future.

**Planner challenges:** The GP search space is large, with 1,662,486 GP total across all land areas that are not urban, frozen or wetlands. The combinatorics grow due to the large number of choices for how and when to view each GP. This includes different combinations of satellites, times, instruments, and viewing angles. The examples in this paper involve 3 satellites, but we plan to add more.

For 3 satellites over a 6-hour period, the constellation sees a total of 1,062,777 GP distributed over 29,700 TP. Each satellite can see an average of 354,259 GP, distributed over an average of 9,900 TP. Each satellite has 2 instruments, each instrument may be used at 62 possible viewing angles. Instruments may be used independently or in various combinations, resulting in an average of 51 command choices/TP with high variance, including a maximum of 108 command choices/TP. Since each satellite has about 10,000 TP (imaging opportunities) with an average of 51 command choices per TP, there are approximately 500,000 options to consider for each satellite.

Another key challenge is that the instrument command search space (defined by orbits and instrument command choices) is different from the science search space (defined by error-dependent utility associated with each plan). Planner output must be a sequence of commands from the command search space, but the value of the plan is defined by the science search space.

Contributions of this work include initial identification, quantification, and qualitative characterization of these two different search spaces (TP choices and GP choices) along with providing an initial integration between them.

Each stripmap image by the Synthetic Aperture Radar (SAR) covers several GPs, that are organized in approximately a 9 km x 9 km grid, that represent the data products of the SMAP mission, a soil moisture flagship mission. While each GP is a single <latitude, longitude> point, every observation captures multiple GP, each with its own model error.

The highly non-linear constraints on slewing and valid instrument combinations present another challenge. We initially started with an MILP formulation but complicated constraints in our time-sliced state model formulation proved challenging. We switched to this constraint processing approach but may revisit the MILP formulation.

### Solution

**Preprocessing:** We start with a preprocessing phase to distill disparate raw data sources into streamlined planner input. The preprocessor prepares the data for the planner by producing files which are structured specifically to define the planner's search space. Preprocessing consolidates, reformats, and compresses the following heterogeneous raw input data sources.

- <u>GP definition</u> file: The planner reads in a file defining 1,662,486 GP along with their biome types (e.g., forest, grasslands, shrubbery, ocean).
- <u>Payload access time</u> files specify the times when each payload for each satellite can view each GP, based on satellite orbits. This file contains tuples of the form: <TP, instrument, viewAngle, GPs>, where GPs is the list of GP in the field of view for the given instrument and viewing angle at the given time TP.
- <u>Rain</u> files specify the GP where rain was recently observed, along with where and when rain is predicted.
- <u>Saturation</u> files specify GP which are already saturated with water.
- <u>Measurement</u> error files specify measurement errors associated with each combination of instruments and viewing angles, for each biome type, as a tuple: <inst1, viewAngle1, inst2, viewAngle2, biomeType>.
- <u>Slew</u> file specifies time and energy required to slew (maneuver) between all viewing angle combinations.

Preprocessing removes redundancies from the raw inputs which are produced by multiple independent and heterogeneous systems and removes data from the raw inputs which are not required by the planner. A key part of pre-processing is a step called choice flattening which eliminates redundancies in the raw input. This eliminates about 66% of the initial (redundant) choices found in the raw input data, which specifies each GP's command choices independently, rather than grouping the commands by TP.

Preprocessing produces multiple files, defining two search spaces which must be integrated during planning (Figure 2). There is a separate TP choice file for each satellite (because each satellite covers GP at different times, and each executes an independent command sequence), but a single GP choice file for the whole constellation.

**Timepoint (TP) choices: Command choices and times for viewing each GP (per satellite)**
- Command search space
- One file for each satellite

| Time | cmd choices | GP covered by choice |
|------|-------------|----------------------|
| 1311: | L.32: | [3165] |
|       | L.34: | [3445, 3446] |
|       | P.32: | [3165] |
|       | P.33: | [3165] |
|       | P.34: | [3445, 3446] |
|       | P.35: | [3445, 3446] |
|       | L.32 & P.32: | [3165] |

**Ground Position (GP) choices:**
- When & how to view each GP
- Science-return search space used in local heuristics
- Measurement error depends on GP biome-type (shrub, forest, baren)
- One file for whole constellation

| GP | satellite | Time | cmd choices | measurement error |
|----|-----------|------|-------------|-------------------|
| 3165: | 1 | 1311 | L.32 | .038 |
|       | 1 | 1311 | P.32 | .017 |
|       | 1 | 1311 | L.32 & P.32 | .003 |
|       | 2 | 1259 | L.33 | .032 |
|       | 2 | 1259 | P.33 | .005 |
|       | 2 | 1259 | L.33 & P.33 | .024 |

**Figure 2: Preprocessing produces TP choice and GP choice files which are the primary planner inputs**

Figure 2 shows the two file types produced by preprocessing which are the planner inputs. The TP choice file (top) defines the timepoints (TP) when some GP is visible. There is one TP choice file for each satellite because each satellite covers GP at different times, and each executes an independent command sequence. The Time column identifies unique TPs when the satellite can see at least one GP. Each TP corresponds to a decision variable in the planner's search space, and the command choices for that TP correspond to that variable's domain. The planner decides which command to choose at each TP for each satellite.

In the TP choices file, each TP maps to a set of command choices, and each command is associated with a set of GP which will be covered by that observation. Command choices are denoted as <instrument(s), viewingAngle>. For example, L.32 means L-band at viewing angle 32. Note that a single command choice may actually involve two observations. The last command choice 'L33 & P.33' means both instruments (L and P) will take an image at angle 33 at the same time. Note that the planner may choose either L.33 or P.33 or both. All three will cover GP 3165, but with different measurement errors. The planner must choose a command for every timepoint when there is something to look at. This is the primary search space for the planner. It starts at the first TP and marches chronologically forward until it fills all available timepoints.

The other file is the GP choice file (bottom), which defines all the choices for how to look at a specific GP. This is a GP-centric view which specifies which command choices are better from each GP's perspective. These GP choices are used by the planner for local heuristics to sort the command choices at each TP choice. The local heuristics rank the command choices for each GP in order of increasing measurement error. These GP choices define different times and commands for viewing the same GP for all satellites. Each GP choice is associated with a different instrument error (which depends on that GP's biome type). These choices include all combinations of access times, instruments, and viewing angles. Note that there are three choices to view GP 3165 at time 1311 at angle 32 by satellite 1, and another 3 choices to view it at time 1259 at angle 33 by satellite 2. Each of these choices has a different measurement error.

**Heuristically Guided Constraint Optimization**

The planner creates an observation plan for a team of satellites, each with two instruments. The planner output is a sequence of time-indexed commands for both instruments, for each satellite. There are 3 command types:

- `TakeImage(<instrument1,viewAngle1> <instrument2,viewAngle2>)`. If both instruments are used at the same TP, they must point at the same angle, but the planner may choose to use the instruments at different times (within 2 hours) in which case the angles may be different, and the planner may choose which instrument to use first. Each command takes 3 seconds to execute.
- `SlewToAngle(fromAngle, toAngle)`: This command takes a variable amount of time to execute depending on the initial and target angles.
- `Idle()`: Turn off the instruments to save energy.

**Plan for satellite 1:**

| Time | Command |
|------|---------|
| [2-4] | P.48 |
| [5-14] | Idle |
| [15-17] | L.48 |
| [18-36] | Idle |
| [37-40] | Slew |
| [41-43] | L.44 |
| [44-45] | Slew |
| [46-48] | P.45 |

**Figure 3: Example planner output for one satellite**

Figure 3 shows an example of planner output for one of the satellites. This example says: From time 2 through 4, take an observation using the P-band at angle 48. Then remain idle for 10 seconds from time 5 through 14. Then use the L-band at the same angle 48 from time 15 through 17. Then there is a 23 second gap until the next observation at time 41. The next observation is at a different viewing angle, so slewing is required, taking 4 seconds of that gap, ending at or before time 40, so the new observation may start at time 41.

Planner Design: We are solving this as a Constraint Optimization Problem (COP) (Dechter 2003), which is defined generically as:
- Set X of variables $\{x_i, .. x_n\}$
- Set D of variable domains $\{d_i, .. d_n\}$ for each variable
- Set C of constraints on legal variable combinations
- Satisfiability requirement: Find a consistent set of variable assignments for all variables for hard constraints
- Optimization: maximize the preference rewards

Our specific DSHIELD COP is defined as:

Problem: Assign commands for every satellite for every TP when it can observe a GP. The TP when a satellite can observe a GP is called an *access time*. There are too many GP to observe them all so this is inherently an optimization problem. The objective is to reduce the average model error by observing as many high-error GP as possible.

Given Inputs from preprocessing (from files in figure 2):
Let G = the set of all GP.
Let $g_i \in$ G = the *ith* GP.
Let $e_{i,t}$ = the model error for $g_i$ at *t*.
Let $m_{i,c}$ = the measurement error for observing $g_i$ with command *c*.
$d_t^s$ (defined formally as the domain for variable $x_t^s$) is the set of commands available for sat s at time t.
Let $v_{s,c,t}$ = the set of all GP visible to sat *s* with command *c* at time *t*.

Note that model error $e_{i,t}$ is a function of time, and increases with time as prior data gets stale, and it increases with rain. When a new observation of $g_i$ is made, the model error $e_{i,t}$ is set to the measurement error $m_{i,c}$.

Decision Variables: We define a set of decision variables $x_t^s$, each representing the command choice for sat s at time t. $\forall\, t \in T^s$, $T^s$ = {All access times (TP) for sat s}.

Variable Domains: We define a set of variable domains $d_t^s$ representing command choices for each $x_t^s$. The domain of choices for $x_t^s$ is the set of all command options for sat s at time t. $d_t^s \in$ {(<instrument1, viewAngle1>, <instrument2, viewAngle2>)}

In our example scenario, there are approximately 10,000 variables per satellite corresponding to the 10,000 TP per satellite, and each variable has an average of 50 command choices in its variable domain.

Our hard constraints include the 3-second image lock, path-dependent slew time, and energy budget. Soft constraints (preferences) include (1) maximizing model improvement, (2) maximizing the # of high-priority GP observed (primary preference) and (3) minimizing energy consumption.

**Objective**
The objective is to: *maximize reduction of model error.* First, we define the *gpReward*, $r_{i,c,t}$, which is the reward for viewing $g_i$ with command *c* at time *t*. This is the difference between the prior error and the measurement error.

$$r_{i,c,t} = e_{i,t-1} - m_{i,c} \qquad (1)$$

If $r_{i,c,t} < 0$ then taking an image with command c would increase model error for $g_i$. Observations which increase error will be discarded, so $r_{i,c,t}$ has a lower bound of 0.

Next, we define *cmdReward*(c, $v_{s,c,t}$) as the sum of *gpRewards* for all GP covered by sat s using command c at time t. This is the aggregate reward for including command c in the plan.

$$cmdReward(c, v_{s,c,t}) = \Sigma_i\, r_{i,c,t}\; \forall\, g_i \in v_{s,c,t}\; \forall\, c \in d_t^s \qquad (2)$$

Our objective is to maximize the sum of all *cmdRewards*, for all commands in the plan. Let P = a plan consisting of a list of commands, and $c_n$ be the *n*'th command in plan P.

**maximize** $\Sigma_{c_n \in P}\, cmdReward(c_n)$ (3)

Equation 3 maximizes the sum of all *gpRewards* for all GP covered by all commands in the plan. The quantity being maximized is the *plan score* which we will use later when describing the search procedure:

$$planScore = \Sigma_{c_n \in P}\, cmdReward(c_n) \qquad (4)$$

**Search Space and Procedure**
The search space is a node tree. Each node is a plan consisting of a (possibly partial) set of variable assignments (cmd choices). Each branch/edge in the tree represents a variable assignment. Each node contains:
- Set of decision variables $x_t^s$, each representing the command choice for sat s at time t. $\forall\, t \in T^s$, $T^s$ = {All GP access times (TP) for sat s}
- Set of variable domains $d_t^s$ representing command choices for each $x_t^s$, $\forall\, t \in$ {access times}. The domain of choices for $x_t^s$ is the set of all command options for sat s at time t.

Node state: Each node also contains a 'state' property, which tracks the energy consumption and battery charge and plan score at each plan step (after each command choice). The node state tracks the effects of each planner choice and is represented as a Python dictionary tracking state fluents such the current battery charge level.

---
Root Node variables:
$x_0^1, x_1^1, x_2^1, x_2^2, x_3^1, x_3^2, x_4^2, x_5^2, x_6^1, ...$

---

Figure 4: Decision variables for the root node

Figure 4 shows and example of the decision variables for the root node. $x_t^s$ = the command for sat s at time t. The root node is initialized with variables for every TP for every satellite. The root node variables are sorted chronologically, so all variables for time N precede all variables for times greater than N. There are variables only for the times when the satellite as access to a GP. This example shows there are variables for only sat 1 at times 0, 1, and 6. There are variables for both satellites at times 2 and 3, and variables for only sat 2 at times 4 and 5. This is because those are the only times when the given sat has TP choices. We may choose to solve the variables in any order, but our default is to solve them in chronological order.

```
1.   PlanIt()
2.      rootNode.vars = {x_t^s}
3.      openNodes = {rootNode}
4.      while openNodes:
           // choose plans to expand (beamWidth # of plans)
5.         beam = chooseNodes(openNodes, beamWidth)
6.         for node in beam:
7.            var  = chooseVariable(node)   // choose x_t^s
8.            val  = chooseValue(var)       // choose cmd
9.            child = createChildNode(node, var, val)
              // propagateChoices enforces constraints and
              // update states properties for energy and reward
10.           child.propagateChoices(var,val)
11.           if child.isFeasible() // verify energy is OK
12.              then openNodes.add(child)
```

**Figure 5**: **Search Procedure**

Figure 5 shows the search procedure PlanIt(). The root node is initialized with variables for all TP for all satellites as shown in figure 4. The algorithm iteratively chooses a set of nodes to expand, called the *beam*. The number of nodes in that set is the *beamWidth*. Each node in the beam is then expanded, which means choosing a variable from that node, and then choosing a value for that variable.

PlanIt() is a generic search engine which may be used for different planning problems. The search engine itself knows nothing about GP, TP, model error or slew constraints. All domain specific logic is encapsulated in planner callbacks provided by the applications. The methods **chooseNodes**, **chooseVariable, chooseValue, propagateChoices**, and **isFeasible** (lines 5, 7, 8,10 and 11) use those domain specific callback methods which contain all knowledge about GP and TP.

Choosing a node corresponds to choosing a plan (the command sequence defined by the path from that node to the root). Choosing a variable corresponds to choosing a <sat, TP> pair and choosing the value corresponds to choosing a command for that sat at time TP. We have used this same system to schedule data downlinks by providing a different set of callbacks for these methods. Future work will integrate the observation and downlink planners.

**chooseNodes** (line 5) implements our objective function using a beam search. On each iteration, the algorithm selects the N best nodes to expand, where N is the beam width and *best* is defined by the node's plan score. Each node has a *plan score* (equation 4) defined as the sum of error reduction for all the GP covered by all the commands in the plan up to that point, for all satellites. All open nodes are sorted in decreasing order of plan score.

**chooseVariable** (line 7) chooses variables in chronological TP order, but may be changed in the future.

**chooseValue** (line 8) implements the local heuristics which sort choices at each choice point (TP). The choices are instrument commands which may be sorted in various ways. We have explored a range of local heuristics, including the following discussed in this paper:

errReduction (prior error - measurement error): commands are sorted in decreasing order of the sum of error reduction for all GP covered by the command. This is the local equivalent of the global objective heuristic, ranking command choices by the same *gpReward* metric that the global objective uses in chooseNodes (equation 1). In this local heuristic version command choices for each variable, $x_t^s$, are sorted in decreasing order of the sum of reduction in error for all GP covered by the command. This is the *gpReward*, $r_{i,c,t}$, shown in equation 1 and the choices are sorted in order of decreasing *cmdReward* (equation 2). This shows how we integrate the TP and GP search spaces.

gpRankedChoice (qualitative measurement error): This is a GP-centric heuristic based on each GP's preference of the best way to view it. Command choices for each TP are sorted based on a form of ranked choice voting in the GP space. Command choices are ranked by the collective set of GP covered by all commands choices for that TP.

First, each GP ranks all its viewing choices across all TP and all satellites, based on measurement error. Rank 1 means the best viewing option (minimum error), and rank 2 is second best, etc. Let $C_i$ = the list of all commands for viewing GP $g_i$, sorted in order of increasing measurement error $m_{i,c}$, so the command with the least measurement error is first. For this heuristic we redefine *gpReward*, $r_{i,c,t}$. Instead of equation 1: $r_{i,c,t} = (|C_i| - c\text{'s position in } C_i) / |C_i|$

Next, the choices for variable $x_t^s$ are sorted so the commands ranked highest by more GP, out of all GP covered by any command at that TP, are chosen first. The choices are sorted in order of decreasing *cmdReward* (equation 2), the same as they were for the errReduction heuristic above, but now the cmdReward is based on a different gpReward calculation. This is another example of how we are integrating the GP search space into the TP search space.

gpCount (no error model): Commands are sorted in decreasing order of the # of GP covered by each command. This is a greedy heuristic which collects as many GP as possible as quickly as possible without regard to model error. We ignore the gpReward, and redefine *cmdReward*(c, $v_{s,c,t}$) to be the # of GP covered by command c (instead of equation 2). At each TP, commands c are sorted in order of decreasing *cmdReward*(c, $v_{s,c,t}$) = $|v_{s,c,t}|$ = the number of GP visible to sat *s* using command *c* at time *t*. This provides an upper bound on # of GP covered and lower bound on our objective score (plan score).

Constraint handlers: Constraints are enforced through *choice propagation* which uses *forward checking* (Russell and Norvig 2021) after each choice to remove any future variables which are inconsistent with it. For example, the 3-second image lock is implemented as follows: when a decision is made to start taking an image at tick 10, then all choices for timepoints 11 and 12 are removed so nothing else will be scheduled during that 3-second hold. Choice propagation is also used to enforce constraints for slew time and duplicate observations. All constraints are implemented by the `propagateChoices` method (line 10).

Constraints are applied only to variables for the same satellite, except for the duplicate observation constraint which is applied to all variables for all satellites. This means the image lock constraint for satellite 1, requires that only satellite 1 hold its position. On the other hand, if satellite 1 observes GP 10, then satellite 2 is constrained to **not** observe GP 10. The following choice propagation example shows how it is used to remove duplicate GP observations from future variables.

(a) $[x_{25}^1: \{\text{L.32: [123]},$
  L.33: [436349, 436350, 436351],
  P.32: [436350, 436351, 436352]]$

(b) $[\cancel{x_{36}^1: \{\text{P.42: [253]}\}}]$

Figure 6: Choice propagation examples

Figure 6 shows two examples of choice propagation to enforce the no duplicates constraint. Case (a) shows the command choices for $x_{25}^1$, satellite 1 at TP 25. After GP 123 is observed, it is removed from the list of GP covered for every choice for all future variables. In this example, that was the only GP covered by command choice L.32, so the command L.32 is removed from the domain of choices for variable $x_{25}^1$. Case (b) shows that when the last choice is removed from a variable's domain (producing an empty domain), then the variable is removed from the node. In this case, after GP 253 is observed, it's removed from the GP list for command P.42, leaving an empty GP list. The command P.42 is removed from the domain for variable $x_{36}^1$, leaving an empty variable domain, so $x_{36}^1$ is removed from the list of open variables. This is different from a pure CSP system where all variables must receive a valid assignment, and a variable with an empty domain indicates infeasibility. In our case, choice propagation removes variables which have no valid assignments based on the path dependencies of the current plan (node). Choice propagation also updates the energy model and the plan score for each node.

**Multiple-pass planning:** We group the full set of GP into priority cohorts based on model error. First priority are rainy GP, where there has been recent or predicted rain. Second priority are GP where there has not been recent or predicted rain. Both of those groups exclude GP where the ground has already been saturated, which are a third priority. On the first pass, the planner schedules as many first-priority (rainy) GP as can fit in the 6-hour plan horizon, then it 'back-fills' the gaps in that plan with non-rainy GP.

This multi-pass approach brings complications. When filling a gap, the planner must enforce the slew constraints to splice the new observations into the existing plan. It must slew from the last high-priority viewing angle at the start of the gap and it must slew to the next planned viewing angle at the end of the gap. Additionally, we cannot propagate the state of battery charge forward in time, because the planner is filling in parts of the plan (gaps) in non-chronological order.

## Experiment Results

Preliminary experiment results are presented below for our example with 3 satellites and a 6-hour planning horizon. All GP are initialized with model error values provided by the Soil Moisture Model. The initial error/GP is 0.0161 (1.61 %) averaged over the entire set of 1.66 million GP.

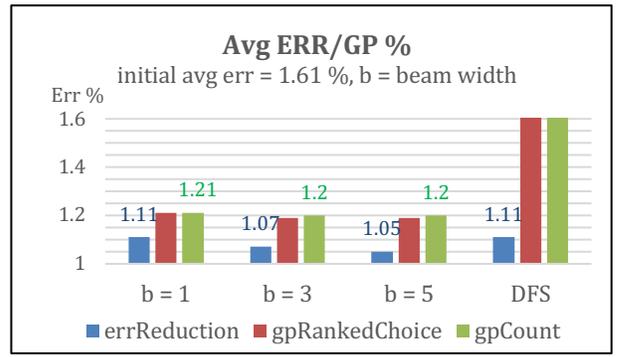

Figure 7: errReduction has best avg err/GP

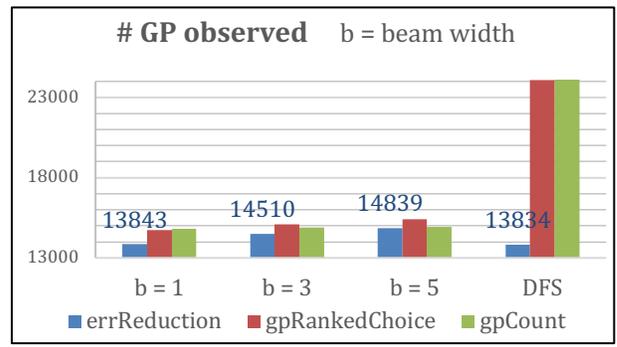

Figure 8: gpCount covers the most GP

Figures 7, 8 and 9 show preliminary experiment results comparing three local heuristics, which are variants of the `chooseValue` method (line 8). Figure 7 shows the average error % per observed GP, figure 8 shows the total # of observed GP, and figure 9 shows how the # of search nodes created scales relative to beam width. We ran the same experiments with beamwidths = 1, 3 and 5, denoted as b in the charts.

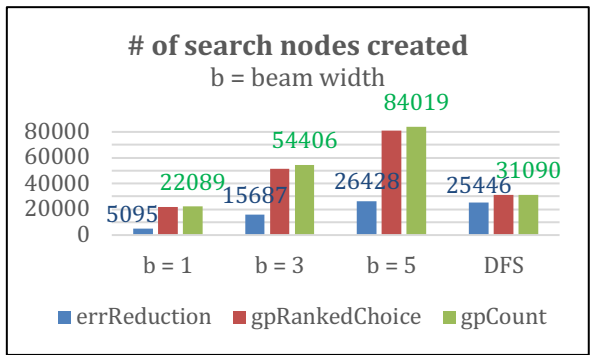

**Figure 9: errReduction & DFS scale linearly with b**

The three heuristics (previously described) are *errReduction*, *gpRankedChoice* and *gpCount*. The group marked DFS on the right of each chart are results when we replace the objective function (equation 3) with a simple depth first search, by changing the definition of the global heuristic `chooseNodes` (line 5). Instead of sorting nodes by error reduction, the list of open nodes is treated as a LIFO stack. The last node added is the first expanded. Comparing DFS results with the others shows how the global heuristic affects the results. The DFS tests used a beamwidth of 5. This means on each call to `chooseNodes`, the 5 most recently added children would be selected.

**Analysis:** These initial results clearly show the inherent trade between quantity and quality, specifically maximizing the # of GP observed vs. maximizing the reduction in model error. Figure 7 shows errReduction always outperforms the others in quality (reducing error) while figure 8 shows gpCount is best for quantity (observing more GP). When using the global objective heuristic (vs. DFS), results generally improve with increased beamwidth.

Figure 9 shows that the # of nodes scales linearly with b only for errReduction (except for DFS group). The poor scaling for gpRankedChoice and gpCount is evidence of tension between the local heuristic (chooseValue) and the global heuristics (chooseNodes), which are conflicting. With errReduction, the local and global heuristics use the same gpReward metric (equation 1), so the heuristics don't conflict. With the other heuristics, more nodes are created because those local heuristics makes choices based on a different gpReward metric, which don't align with the global objective. The global objective acts like a governor, and repeatedly backtracks to correct the diversion. The global objective to maximize error reduction clamps down on the other heuristics from chasing their local preferences. Note that all heuristics in the DFS group scaled linearly with beamwidth because there's no conflict between the global and local heuristics, so the local heuristic is untethered from the global heuristic's veto power.

Other observations (not shown in charts) include: All mixes of heuristic and beamwidth take roughly the same number of images (about 5075). This corresponds to the # of commands in the plan and corresponds to the makespan. We also observed that beam width has minimal effect on the solution with DFS because without the global heuristic, a beamwidth of 1 and a beamwidth of 5 produce similar locally greedy solutions. More nodes are generated with larger beam width, but not exponentially more.

**Future Work:**

Global heuristics and search strategies: We plan to test variations of the `chooseNode` method beyond the objective-based and depth first methods used in this paper including a branch-and-bound algorithm (Dechter 2003) which prunes suboptimal nodes. For example, if two nodes have similar plan scores but one got there earlier in the plan horizon (and thus has more time remaining in the horizon to make more observations), then `chooseNodes` may prune the suboptimal node and backtrack to the best previous node. We may also modify `chooseNodes` to implement Monte Carlo rollouts or other randomized search methods.

More local heuristics: We are exploring other variants for our local heuristics.

GP-choice search space: We plan to explore planning observations based on GP choices as the primary search space rather than using TP choices as our search space. Intuitively this means solving the most important GP first rather than solving each TP in chronological order. This can be implemented by changing the `chooseVariable` method to select variables which cover GP with the highest error in non-chronological order.

MILP formulation: We plan to revisit an MILP formulation, with the increased insight and better understanding of the problem gained by developing this COP solution.

Downlink planning: We previously used this same COP system to implement a downlink planner. That planner tracked two priority levels of collected data and scheduled timeslots on ground stations to download the data in priority order. It used the same algorithm as Figure 5 but was provided with different callbacks for `chooseVariable` and `chooseValue`. We plan to integrate that downlink planner with the observation planner presented here.

Constellation and scenario extensions: We will add more satellites, and more instruments to each satellite (L band radiometer, P band reflectometer, L band reflectometer) to improve soil moisture retrievals. These instruments will add additional choices to optimize energy consumption, albeit an order of magnitude less than the radar.

## Acknowledgements

This work has been funded by the NASA Earth Science Technology Office's Advanced Information Systems Technology program. We acknowledge the important contributions of every member of the D-SHIELD. Thank you to Samantha Niemoeller for very helpful feedback.

# References

Augenstein, S., Estanislao, A., Guere, E., and Blaes, S. 2016. "Optimal scheduling of a constellation of earth-imaging satellites, for maximal data throughput and efficient human management".

Barnsley, M., Settle, J., Cutter, M., Lobb, D., and Teston, F. 2004. "The PROBA/CHRIS mission: A low-cost smallsat for hyperspectral multiangle observations of the earth surface and atmosphere," Geosci. Remote Sens. IEEE Trans. On, vol. 42, no. 7.

Chien, S.; Doubleday, J.; Mclaren, D.; Davies, A.; Tran, D.; Tanpipat, V.; Akaakara, S.; Ratanasuwan, A.; and Mandl, D. 2011. Space-based Sensorweb Monitoring of Wildfires in Thailand. In International Geoscience and Remote Sensing Symposium (IGARSS 2011), Vancouver, BC, July 2011.

Chien, S.; Doubleday, J.; Thompson, D. R.; Wagstaff, K.; Bellardo, J.; Francis, C.; Baumgarten, E.; Williams, A.; Yee, E.; Stanton, E.; and Piug-Suari, J. 2016. Onboard Autonomy on the Intelligent Payload EXperiment (IPEX) CubeSat Mission. Journal of Aerospace Information Systems (JAIS). April 2016.

Chien, S., Mclaren, D., Doubleday, J., Tran, D., Tanpipat, V., and Chitradon, R. 2019. Using Taskable Remote Sensing in a Sensor Web for Thailand Flood Monitoring. Journal of Aerospace Information Systems (JAIS), 16(3): 107-119. 2019.

Chien, S. A.; Davies, A. G.; Doubleday, J.; Tran, D. Q.; Mclaren, D.; Chi, W.; and Maillard, A. 2020. Automated Volcano Monitoring Using Multiple Space and Ground Sensors. Journal of Aerospace Information Systems (JAIS), 17:4: 214-228. 2020.

Dechter. R, 2003. Constraint Processing. Morgan Kaufmann Publishers Inc., San Francisco, CA, USA.

Doubleday J. 2015. "Autonomy for remote sensing — Experiences from the IPEX CubeSat," in 2015 IEEE International Geoscience and Remote Sensing Symposium (IGARSS), Jul. 2015, pp. 5308–5311

Fruth, Thomas, Christoph Lenzen, Elke Gross, and Falk Mrowka. 2019. "The EnMAP Mission Planning System." In Space Operations: Inspiring Humankind's Future, pp. 455-473. Springer, Cham, 2019

Jian, L., and Cheng, W. 2008. "Resource planning and scheduling of payload for satellite with genetic particles swarm optimization," in Evolutionary Computation, 2008. CEC 2008.(IEEE World Congress on Computational Intelligence). IEEE Congress on, 2008, pp. 199–20

Krieger, G., Alberto, M.,, Hauke F., Hajnsek, I., Werner, M., Younis, M., Zink, M. "TanDEM-X: A satellite formation for high-resolution SAR interferometry." IEEE Transactions on Geoscience and Remote Sensing 45, no. 11 (2007): 3317-3341.

Linnabary, R., O'Brien, A., Smith, G., Ball, C., and Johnson, T. 2019. "Open Source Software For Simulating Collaborative Networks Of Autonomous Adaptive Sensors," in IGARSS 2019-2019 IEEE International Geoscience and Remote Sensing Symposium, pp. 5301–5304.

Mandl, D., Frye, S., Goldberg, M., Habib, S., and Talabac, S. 2006. "Sensor webs: Where they are today and what are the future needs?," in Dependability and Security in Sensor Networks and Systems, 2006. DSSNS 2006. Second IEEE Workshop on, 2006

Le Moigne, J., Little, M., and Cole, M. 2019. "New Observing Strategy (NOS) for Future Earth Science Missions," IGARSS 2019 - 2019 IEEE International Geoscience and Remote Sensing Symposium, 2019.

Nag S., et al., 2019. "Autonomous Scheduling of Agile Spacecraft Constellations with Delay Tolerant Networking for Reactive Imaging," presented at the International Conference on Planning and Scheduling (ICAPS) SPARK Applications Workshop, Berkeley, California, U.S.A., 2019

Nag, S., Moghaddam, M., Selva, D., Frank, J, Ravindra, V., Levinson, R., Azemati, A., Aguilar, A., Li, A., Akbar, R., 2020. D-Shield: Distributed Spacecraft with Heuristic Intelligence to Enable Logistical Decisions, Proc. of the 2020 IEEE International Geoscience and Remote Sensing Symposium.

Nag, S., Moghaddam, M., Selva, D., Frank, J., Ravindra, V., Levinson, R., Azemati, A., Gorr, B., Li, A., Akbar, R. 2021. "Soil Moisture Monitoring using Autonomous and Distributed Spacecraft (D-SHIELD)", IEEE International Geoscience and Remote Sensing Symposium, Virtual, July 2021

Russell, S., Norvig, P. 2020. Artificial Intelligence: A Modern Approach. Fourth Edition. Pearson.

Werninghaus, Rolf, and Stefan Buckreuss. 2009. "The TerraSAR-X mission and system design." IEEE Transactions on Geoscience and Remote Sensing 48.2 (2009): 606-614.

Wörle, Maria Theresia, Christoph Lenzen, Tobias Göttfert, Andreas Spörl, Boris Grishechkin, Falk Mrowka, and Martin Wickler. 2014. "The Incremental Planning System GSOC's Next Generation Mission Planning Framework." In SpaceOps 2014 Conference, p. 1785. 2014.